\documentclass{article}

\usepackage{PRIMEarxiv}

\usepackage[utf8]{inputenc} 
\usepackage[T1]{fontenc}    
\usepackage{hyperref}       
\usepackage{url}            
\usepackage{booktabs}       
\usepackage{amsfonts}       
\usepackage{nicefrac}       
\usepackage{microtype}      
\usepackage{lipsum}
\usepackage{fancyhdr}       
\usepackage{graphicx}       
\graphicspath{{media/}}     
\usepackage{setspace}
\usepackage{algorithm}
\usepackage{algorithmic}
\usepackage{amsfonts}
\usepackage{amsmath}
\usepackage{booktabs}
\title{Fast Adaptation with Kernel and \\ Gradient based Meta Learning
}

\author{
  JuneYoung Park \\
  Opt-AI\\
  Department of Convergence Healthcare Medicine \\
  Ajou University \\
  \texttt{jyoung.park@opt-ai.kr} \\
   \And
  MinJae Kang \\
  School of Electronic and Electrical Engineering\\
  Sungkyunkwan University \\
  \texttt{kminjae618@gmail.com} \\
}

\begin{document}
\maketitle

\begin{abstract}
Model Agnostic Meta Learning or MAML has become the standard for few-shot learning as a meta-learning problem. MAML is simple and can be applied to any model, as its name suggests. However, it often suffers from instability and computational inefficiency during both training and inference times. In this paper, we propose two algorithms to improve both the inner and outer loops of MAML, then pose an important question about what 'meta' learning truly is. Our first algorithm redefines the optimization problem in the function space to update the model using closed-form solutions instead of optimizing parameters through multiple gradient steps in the inner loop. In the outer loop, the second algorithm adjusts the learning of the meta-learner by assigning weights to the losses from each task of the inner loop. This method optimizes convergence during both the training and inference stages of MAML. In conclusion, our algorithms offer a new perspective on meta-learning and make significant discoveries in both theory and experiments. This research suggests a more efficient approach to few-shot learning and fast task adaptation compared to existing methods. Furthermore, it lays the foundation for establishing a new paradigm in meta-learning.
\end{abstract}

\vspace{0.2cm}
\vspace{0.2cm}

\section{Introduction}
Deep learning has transformed artificial intelligence by automating the process of learning features from data, which used to require manual input and domain-specific knowledge. This advancement has led to significant improvements in areas like image recognition and natural language processing. However, achieving these results usually demands large amounts of data and complex models with millions, or even billions, of parameters. Training such models requires substantial computational power and resources, which can be challenging to obtain, especially in situations where data or computational resources are limited.

To overcome these challenges, researchers have explored methods to create models that can generalize across various tasks. One popular method involves using large pre-trained models that can be fine-tuned for specific tasks. Although this approach has shown promise, it often requires extensive computational resources and may not perform well if not carefully adjusted for the target task.

Meta-learning, often called ``learning to learn," has emerged as a promising solution to these limitations. Meta-learning enables models to quickly and efficiently adapt to new tasks, even with limited data. For example, in few-shot learning scenarios, where a model must learn from just a few examples, meta-learning techniques have shown to be particularly effective.

Among the different meta-learning approaches, Model-Agnostic Meta-Learning (MAML) \cite{pmlr-v70-finn17a} is one of the most influential. MAML is an optimization-based technique that helps models learn a good starting point (or initialization) that can be quickly adapted to new tasks with minimal data. MAML operates through two main processes: the inner loop, which focuses on adapting the model to a specific task, and the outer loop, which optimizes the learning process across multiple tasks. The interaction between these two loops allows the model to find an initialization that enables fast learning on new tasks.

However, MAML is not without its limitations. The repeated optimization in both the inner and outer loops can be computationally expensive, leading to longer training times, especially with large datasets or complex models. Additionally, the performance of MAML depends on how similar the tasks are to each other. If the tasks are very different, MAML may not work as well because it relies on similarities between tasks to generalize effectively. \cite{raghu2019rapid}

To address these issues, recent research has proposed refinements to MAML, such as reducing the computational burden by improving the optimization process and developing techniques that better capture the similarities between tasks.

In this paper, we identify key challenges in MAML and introduce two improved meta-learning techniques. The first method redefines the optimization process by using closed-form solutions instead of the traditional gradient steps in the inner loop, making the adaptation process more efficient. The second method enhances the outer loop by dynamically adjusting the importance of each task's contribution, leading to better performance during both training and application.

These improvements not only address the computational and stability issues associated with MAML but also offer new insights into the underlying principles of meta-learning. Our work aims to advance meta-learning research by providing a more efficient approach to few-shot learning and paving the way for a new paradigm in this field.

\vspace{0.4cm}
\section{Related Works}
\label{sec:headings}


\subsection{Meta-Learning}
Meta-Learning or ``Learning to Learn" is a rapidly growing field in machine learning that focuses on developing algorithms capable of learning effectively from limited data. According to \cite{thrun1998learning}, the key idea of meta-learning is to create models that can identify patterns from just a few examples. Currently, meta-learning models are classified into four main approaches: black-box adaptation, optimization-based, non-parametric, and Bayesian meta-learning. 

One significant advancement in this field is the introduction of episode-based training, inspired by the Matching Network. According to \cite{chen2019closer}, this approach involves two main stages: meta-training and meta-testing. A notable framework in this area is MAML (Model-Agnostic Meta-Learning), introduced by \cite{pmlr-v70-finn17a}. MAML uses a bi-level loop optimization process, where the inner loop adapts meta-parameters to task-specific parameters, and the outer loop minimizes the overall loss. While MAML has shown promising results, the inefficiency in calculating meta-gradients remains a challenge. Current research in meta-learning is focused on solving these efficiency issues and improving models ability to learn from limited data across various tasks.

\subsection{Few-Shot Classification}
Few-Shot Classification is an important concept in Meta-Learning, referring to the ability of a model to quickly learn new tasks based on very few training examples. Typically, deep learning models require a large amount of data and long training times, but Few-shot Learning aims to overcome these limitations. This approach is especially useful in fields where collecting large datasets is challenging, such as medical imaging, rare language datasets, and personalized systems. 

The main idea behind Few-shot Learning is to enable a model to learn patterns and knowledge that can be generalized across different tasks, so it can quickly adapt to new tasks. This process usually involves two main stages. The first stage is the meta-learning phase, where the model learns broad knowledge by training on a variety of tasks. The second stage is the fine-tuning phase, where the model is adjusted to perform specific tasks using only a few new samples. 

One commonly used technique in this process is simulation or episode learning. In the meta-learning phase, the model is trained through numerous ``episodes." Each episode simulates a new task, helping the model learn to adapt quickly. For example, in an image classification task, each episode might include a few image classes and their corresponding images. Through these episodes, the model develops the ability to quickly learn to classify new image classes. Another important approach is the Prototype Network. This method involves learning a ``prototype" for each class, which represents the typical example of that class. When a new sample is presented, the model classifies it by comparing it to the closest prototype. This approach allows the model to maintain high classification performance even with a small number of samples. 

In summary, Few-shot Learning is a key concept in Meta-Learning that allows models to perform well even with limited data. It shows great potential for innovative applications across various fields, significantly expanding the usefulness of deep learning models in environments where data is scarce.

\subsection{Domain Generalization}
Domain Generalization is one of the key challenge in meta-learning. Domain generalization aims to ensure that a model performs well not only in the domains it was trained on but also in new, unseen domains. This is crucial for real-world applications where models need to maintain consistent performance even in environments they haven't encountered before. Domain generalization is closely related to domain adaptation, but the two concepts differ in important ways. While domain adaptation focuses on optimizing a model for a specific target domain, domain generalization seeks to improve a model's performance across a range of potential domains without specifying a particular target domain. This requires the model to learn generalized representations that can be applied to a broader set of domains. 

Various methods have been proposed to achieve domain generalization. Early research primarily focused on learning domain-invariant features. For instance, \cite{muandet2013domain} proposed a method to learn domain-invariant transformations to reduce differences between domains. This approach allows the model to learn common features across various domains, helping it maintain performance in new domains. 

More recently, methods using meta-learning for domain generalization have gained attention. \cite{li2018learning} proposed a model that, after being trained on multiple source domains through meta-learning, could generalize well to new, unseen domains. This approach involves two stages: meta-training and meta-testing. In the meta-training stage, the model learns a generalized meta-model based on tasks from various source domains. During meta-testing, the model is evaluated on a new domain, and the meta-model's learned patterns are applied to achieve high performance in this new environment.

In conclusion, research on domain generalization through meta-learning offers significant potential for developing models that can perform well in domains they have not been trained on. This approach represents an important and expanding field of study, with wide applicability to real-world problems.


\vspace{0.4cm}
\section{Method}
\label{sec:others}
Meta-learning is a powerful paradigm that allows models to quickly adapt to new tasks with limited data. This approach has gained significant attention in the field of artificial intelligence in recent years. However, existing meta-learning methods have some important limitations. Many of these approaches rely on gradient-based optimization in the parameter space, which can be computationally expensive and slow, especially when dealing with complex tasks or large datasets. Additionally, these methods often require careful tuning of hyperparameters, and they are prone to overfitting when the number of tasks is small or when there is significant variation between tasks. To address these challenges, we propose a new approach that improves the inner-loop optimization in the Gradient-based Meta-Learning's bi-level loops learning process. Specifically, we introduce a method called Adaptive Meta-Learning in Functional Space (AMFS). Unlike traditional methods that operate in the parameter space, AMFS works directly in the functional space, enabling more efficient and generalized adaptation. In AMFS, we use a Radial Basis Function (RBF) kernel, also known as a Gaussian kernel, to compute a closed-form solution for task-specific adaptation. By integrating a composite regularization technique, AMFS overcomes the limitations of traditional meta-learning methods, offering a more robust and effective way to handle diverse tasks.



\subsection{What is Meta-Learning Problem}
The goal of few-shot meta-learning is to create a model that can quickly adapt to new tasks using only a small number of data points and training iterations. To achieve this, the model, or learner, goes through a meta-learning phase where it is trained on a variety of tasks. This training allows the model to quickly adapt to new tasks with just a few examples or attempts. Essentially, in a meta-learning problem, we treat each task as a single training example. The meta-learning framework is designed to help the model adapt to different tasks, which are represented by a distribution called \(p(T)\). In a K-shot learning setting, our goal is to train the model so that it can quickly learn a new task \(T_{i}\), randomly chosen from \(p(T)\), using only minimal information. Specifically, the model must learn using just K examples and the task's evaluation criterion \(\mathcal{L}_{T_{i}}\). The meta-learning process works as follows. First, a task \(T_{i}\) is randomly selected from \(p(T)\), and K training examples are drawn from this task. The model then learns using these K examples and the evaluation criterion \(\mathcal{L}_{T_{i}}\) for \(T_{i}\). After learning, new test samples are drawn from \(T_{i}\) to evaluate the model's performance. Based on the test results, the parameters of the model f are updated, with the test errors from each \(T_{i}\) providing the training signal for the overall meta-learning process. Once meta-learning is complete, we assess the model's ability to generalize by selecting a new, unseen task from \(p(T)\). The model is given only K examples from this new task to learn from, and after training, we measure its performance to evaluate the effectiveness of the meta-learning. It's important to note that the tasks used for meta-testing are different from those used during meta-learning. This ensures that we are truly testing the model's ability to quickly adapt to new tasks.

\begin{algorithm}
\caption{Inner-Loop Framework for Our Algorithm}
\begin{algorithmic}[1]
\REQUIRE $p(T)$: distribution over tasks
\REQUIRE $\lambda, \mu, \gamma$: hyperparameters for regularization
\REQUIRE $\eta$: learning rate for kernel parameters
\STATE Randomly initialize kernel parameters $\theta$
\WHILE{not done}
    \STATE Sample batch of tasks $T_i \sim p(T)$
    \FORALL{$T_i$}
        \STATE Sample $K$ datapoints $D = \{(x^{(j)}, y^{(j)})\}$ from $T_i$
        \STATE Compute kernel matrix $K$ using $k_\theta$
        \STATE Define adapted function: \\ $f^*_{T_i}(x) = \sum_{j} \alpha_j k_\theta(x^{(j)}, x)$
        \STATE Compute closed-form solution: $\alpha = (K + \lambda I)^{-1} y$
        \STATE Sample datapoints $D'_i = \{(x^{(j)}, y^{(j)})\}$ from $T_i$ for meta-update
        \STATE Compute total loss:
        \[
        L_{\text{total}} = \mathcal{L}_{T_{i}}(f_{\theta , i}) - \mu \left\|\nabla_{f}\mathcal{L}_{T_{i}} (f_{\theta , i})\right\|^2 + \gamma I(f^*_{T_i}; D)
        \]
    \ENDFOR
    \STATE Update kernel parameters: $\theta = \theta - \eta \nabla_\theta \sum_i L_{\text{total}}(T_i)$
\ENDWHILE
\end{algorithmic}
\end{algorithm}

\subsection{Adaptive Meta-Learning in Functional Space}
Before explaining the algorithms we propose, let's clarify the terminology. First, the algorithm that improves the inner loop based on kernels is referred to as I-AMFS, while the algorithm that improves the outer loop based on gradient similarity from the inner loop is called O-AMFS. These two algorithms form what we refer to as the AMFS framework.

\subsection{Kernel-Based I-AMFS}
I-AMFS introduces several key innovations that set it apart from traditional meta-learning methods. Unlike conventional approaches that operate in parameter space, I-AMFS works directly in the functional space and uses an RBF (Radial Basis Function) kernel to learn task-specific representations. The RBF kernel is a function that measures the similarity between two data points, effectively capturing nonlinear patterns. One of the major advantages of I-AMFS is that it avoids the repetitive, gradient-based inner loop used in traditional methods. Instead, it calculates a closed-form solution for each task, which greatly reduces computational complexity and enables faster adaptation. This approach not only speeds up the learning process but also simplifies the overall algorithm, making it more robust to variations in task complexity. I-AMFS also incorporates a composite regularization strategy that combines gradient norm regularization and information-theoretic regularization. This dual regularization approach helps prevent overfitting and promotes better generalization, ensuring that the model can quickly adapt to new tasks while avoiding becoming too specialized. Additionally, I-AMFS dynamically learns the kernel parameters during the meta-learning process, providing flexible representations that better match the characteristics of each task. This allows I-AMFS to achieve better adaptability across a wide range of tasks. The objective function of I-AMFS is designed to balance quick adaptation with strong generalization. It is carefully constructed to ensure that the model can adapt swiftly to new tasks while maintaining robust performance across different scenarios. This objective function is defined as follows:
\begin{spacing}{1.5}
    \begin{equation}
    \min_{\theta} \mathbb{E}_{T_{i}\sim p(T)}\{\mathcal{L}_{T_{i}}(f_{\theta , i}) - \mu \Vert \nabla_{f}\mathcal{L}_{T_{i}} (f_{\theta , i}) \Vert^2 + \gamma I (f^{*}_{\theta , i};D_{i})\}
    \end{equation}  
\end{spacing}

Each term in this equation has a specific meaning. $\mathcal{L}_{T_{i}}(f_{\theta , i})$ represents the empirical loss for task $T_{i}$. This measures how well the adapted function $f_{\theta , i}$ fits the data for that specific task. The term $\Vert \nabla_{f}\mathcal{L}_{T_{i}} (f_{\theta , i}) \Vert^2$ is a gradient norm regularization term. This penalizes cases where the gradient of the function $f_{\theta , i}$, defined by an RBF kernel, is too large. By doing this, AMFS is encouraged to find smooth and stable solutions, which helps improve generalization across different tasks. The term $I (f^{*}_{\theta , i};D_{i})$ is an information-theoretic regularization term. It controls how much information the adapted function captures from the training data $D_{i}$, helping to prevent overfitting. This term ensures that the model doesn’t become overly specialized to the tasks it has learned, thereby improving its ability to generalize to new tasks. Each of these components plays an important role in achieving the goals of I-AMFS. 

First, gradient norm regularization ensures that the adapted function is not only accurate but also smooth. A smooth function is less likely to overfit when there is only a small amount of data or when the data is noisy, which is very important in few-shot learning scenarios. Furthermore, a smooth function is more likely to generalize well to new data, making it a key factor in the success of I-AMFS. Information-theoretic regularization controls how much information the model captures from the training data, preventing the model from becoming too specialized to the tasks it has already learned. This helps the model maintain its ability to generalize to new tasks, which is essential in meta-learning.
The closed-form solution for task adaptation in AMFS is derived from the following optimization problem:
\begin{spacing}{1.5}
    \begin{equation}
        f^{*}_{\theta, i} = \arg \min_{f}[\sum (f(x)-y)^2 + \lambda \Vert f \Vert^{2}]
    \end{equation}
\end{spacing}

The solution to the above optimization problem can be expressed as follows:
\begin{spacing}{1.5}
    \begin{equation}
        f^{*}_{\theta, i}(x) = \sum_{j}\alpha_{j}k_{\theta}(x_{j}, x)
    \end{equation}
\end{spacing}

Here, $k_{\theta}(x_{j}, x)$ is the RBF kernel function, which measures the similarity between two points $x_j$ and $x$. The RBF kernel is typically defined as follows:

\begin{spacing}{1.5}
    \begin{equation}
        k_{\theta}(x_{j}, x) = \exp (- \frac{\Vert x_j - x \Vert^2}{2 \sigma^2})
    \end{equation}
\end{spacing}

where $\sigma$ is a hyperparameter that controls the width of the kernel. In the closed-form solution, the coefficient $\alpha_j$ is given as follows:
\begin{spacing}{1.5}
    \begin{equation}
        \alpha = (K_{\theta} + \lambda I)^{-1}y
    \end{equation}
\end{spacing}

Here, $K_{\theta}$ is the kernel matrix calculated using the RBF kernel, and $I$ is the identity matrix. This closed-form solution offers several advantages. In terms of efficiency, gradient-based methods require multiple iterations to converge, but a closed-form solution can be computed directly in a single step. This reduces computational complexity and allows I-AMFS to adapt to new tasks more quickly. In terms of stability, the closed-form solution avoids potential issues like unstable local minima or slow convergence that can occur with gradient descent, enabling the model to adapt to new tasks in a stable and consistent way.

Another unique feature of I-AMFS is its ability to learn the kernel parameters during the meta-learning process. Traditional kernel methods rely on a fixed kernel function, which can limit the model's ability to adapt to different types of tasks. In contrast, I-AMFS dynamically adjusts the kernel parameters $\theta$ as part of the meta-learning process, allowing the model to better tune the underlying representation to the specific characteristics of each task.
If we define the objective function as $J(\theta)$, the meta-optimization in AMFS is performed through the following gradient update:
\begin{spacing}{1.5}
    \begin{equation}
        \theta \leftarrow \theta-\eta \nabla_{\theta}J(\theta)
    \end{equation}
\end{spacing}

This update rule ensures that the kernel parameters are optimized to balance task-specific adaptation and generalization. By learning the kernel parameters, I-AMFS can flexibly adjust the complexity of the model, providing representations that better adapt to a variety of tasks.

In conclusion, I-AMFS addresses the limitations of traditional meta-learning methods by performing the inner loop tasks in function space, leading to fast and efficient few-shot learning. By combining the use of closed-form solutions, complex regularization, and dynamic kernel learning, AMFS offers a robust and efficient inner loop framework for quick and generalizable adaptation across different tasks.

\begin{figure}
    \centering
    \includegraphics[width=10cm]{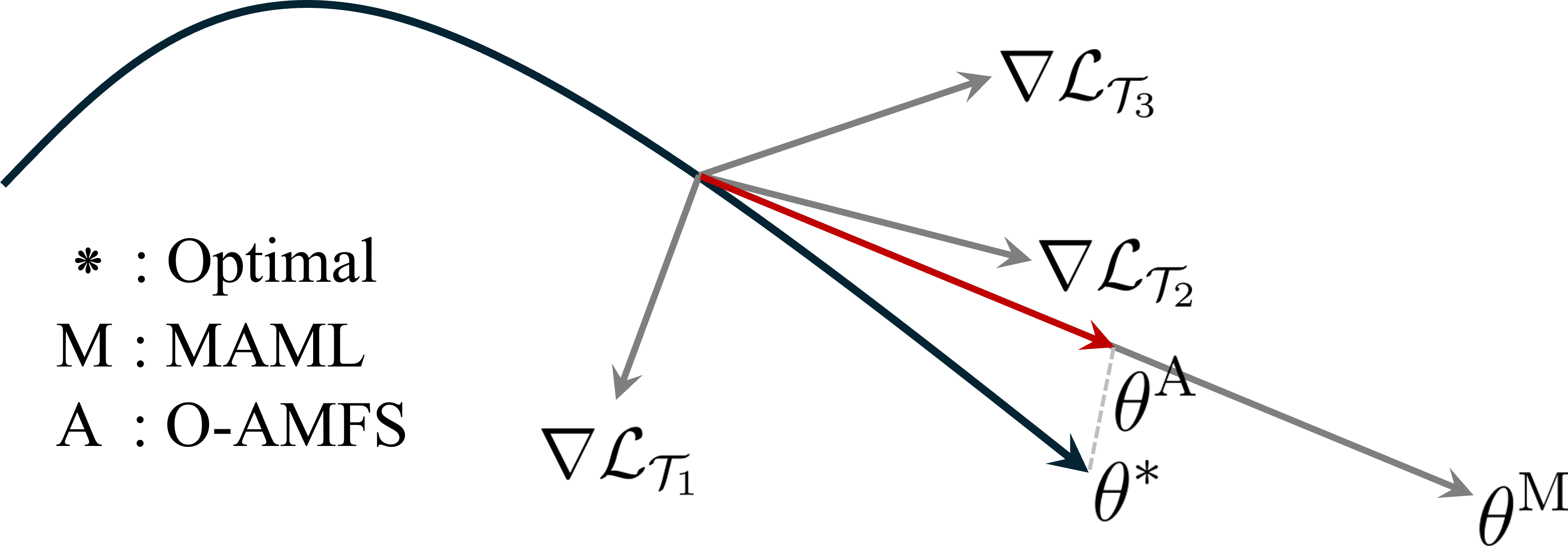}
    \caption{Diagram of our AMFS algorithm in outer loop (O-AMFS), meta-gradient is adaptively determined based on the similarity of the gradients for each task.}
    \label{fig:enter-label}
\end{figure}

\subsection{Gradient-Based O-AMFS}
In the original Model-Agnostic Meta-Learning (MAML) algorithm, the model adapts to specific tasks by updating task-specific parameters, denoted as $\theta_i^\prime$, within the inner loop for each task $\mathcal{T}_i$. These task-specific updates are then used to adjust the meta-parameters $\theta$, which guide the model's performance across all tasks. The meta-parameters are updated as follows:
\begin{spacing}{1.5}
    \begin{equation}
        \theta \leftarrow \theta-\beta\nabla_\theta\sum_{\mathcal T_i\sim p(\mathcal T)}\mathcal L_{\mathcal T_i}(f_{\theta_i^\prime})
    \end{equation}
\end{spacing}

where $\beta$ denotes the meta-learning rate. 

In this formulation, the meta-parameters are updated by simply summing the gradients $\nabla_\theta \mathcal{L}_{\mathcal{T}_i}$ obtained from each task in the inner loop. However, this method can run into problems when the tasks are not similar. When the tasks differ significantly, summing the gradients can lead to poor performance because the updates might conflict with each other. \cite{raghu2019rapid}

To address this issue, we propose a new algorithm called O-AMFS. This algorithm aims to prevent performance degradation caused by differences between tasks by adjusting the meta-parameters based on similarity between the gradients of different
tasks. Specifically, O-AMFS updates the meta-parameters by giving more weight to the gradients of tasks that are more similar to each other as follows:
\begin{spacing}{1.5}
    \begin{equation}
        \nabla_\theta\mathcal L_{\mathcal T_i}(f_{\theta_i^\prime})\leftarrow 1+ \frac{1}{m}\sum_{j\neq i}^m w_{ij}\nabla_\theta\mathcal L_{\mathcal T_i}(f_{\theta_i^\prime})
    \end{equation}
\end{spacing}

Here, $m$ denotes the number of tasks sampled in the inner loop, and the weight $w_{ij}$ is defined as the cosine similarity between the gradients of task $\mathcal{T}_i$ and task $\mathcal{T}_j$:
\begin{spacing}{1.5}
    \begin{equation}
        w_{ij}=\cos(\nabla_\theta\mathcal L_{\mathcal T_i}(f_{\theta_i^\prime}),\nabla_\theta\mathcal L_{\mathcal T_j}(f_{\theta_j^\prime}))
    \end{equation}
\end{spacing}

This approach prevents conflicting updates when the gradient directions of different tasks vary significantly, and emphasizes the gradients of similar tasks, leading to more consistent and effective meta-parameter updates.

The theoretical foundation of this method is based on the idea that, in few-shot meta-learning, each task in the inner loop learns from a subset of the overall data distribution. Therefore, the gradients learned from each task can be used to approximate the optimal direction for the overall data distribution. Under this assumption, the O-AMFS method aggregates gradients by considering their similarity in direction, resulting in meta-parameter updates that more accurately represent the overall data distribution. Through the experiments, we have demonstrated the validity of this assumption, showing that the O-AMFS algorithm indeed contributes to faster convergence and more stable learning.

\begin{algorithm}
\caption{Outer-Loop Framework for O-AMFS}
\begin{algorithmic}[1]
\REQUIRE $p(\mathcal{T})$: distribution over tasks
\REQUIRE $\beta$: learning rate for meta learner
\STATE Randomly initialize meta parameters $\theta$
\WHILE{not done}
    \STATE Sample a batch of tasks $\mathcal{T}_i \sim p(\mathcal{T})$
    \FORALL{$\mathcal{T}_i$}
        \STATE Sample $K$ datapoints $D_i = \{(x_i^{(j)}, y_i^{(j)})\}$ from $\mathcal{T}_i$
        \STATE Compute adapted parameters $\theta_i^\prime$ using gradient descent on task $\mathcal{T}_i$
        \STATE Compute task loss: $\mathcal{L}_{\mathcal{T}_i}(f_{\theta_i^\prime})$
        \STATE Compute weighted gradient for task $\mathcal{T}_i$:
        \[
        \nabla_\theta \mathcal{L}_{\mathcal{T}_i}(f_{\theta_i^\prime}) \leftarrow 1 + \frac{1}{m} \sum_{j \neq i}^m w_{ij} \nabla_\theta \mathcal{L}_{\mathcal{T}_i}(f_{\theta_i^\prime})
        \]
        \STATE where $w_{ij} = \cos(\nabla_\theta \mathcal{L}_{\mathcal{T}_i}(f_{\theta_i^\prime}), \nabla_\theta \mathcal{L}_{\mathcal{T}_j}(f_{\theta_j^\prime}))$
    \ENDFOR
    \STATE Update meta parameters: $\theta \leftarrow \theta - \beta \sum_i \nabla_\theta \mathcal{L}_{\mathcal{T}_i}(f_{\theta_i^\prime})$
\ENDWHILE
\end{algorithmic}
\end{algorithm}

Moreover, this method is similar to the weighted aggregation techniques used in federated learning, where gradients from different clients are combined based on their relevance. This analogy suggests potential future research directions, exploring other ways to aggregate task-specific gradients within the O-AMFS framework.

\vspace{0.4cm}
\section{Experimental Setup}
\subsection{Baseline}
We used the MAML framework as the baseline for our experiments, and all experimental results were obtained by conducting a locally replicating the original implementation following \cite{pmlr-v70-finn17a}.
\subsection{Datasets} We conducted a series of experiments to evaluate the effectiveness of the proposed framework. The experiments primarily include few-shot classification, scenario study, and ablation study. The datasets used for the experiments are Omniglot \cite{lake2011one}, Mini-ImageNet \cite{ravi2016optimization}, FC-100 \cite{oreshkin2018tadam}, and CUB \cite{wah2011caltech}, which are widely used in meta-learning. Through these datasets, we comprehensively evaluate the performance of our methodology across a diverse spectrum of tasks.
\subsection{Environments} We implemented our methodology, as well as MAML and f-MAML, using Python 3.8, PyTorch 1.8.1, and the torchmeta library \cite{deleu2019torchmeta}, ensuring a consistent experimental environment with the use of an A100 GPU.
\subsection{Hyperparameter} In line with \cite{pmlr-v70-finn17a}, we sampled 60,000 episodes for our experiments. Additionally, we used the 4-Convolution architecture, as detailed in \cite{vinyals2016matching}, to apply our parameter update method. The learning rates were set to 0.01 for the inner loop and 0.001 for the outer loop. We also used the same number of gradient steps in the inner loop as in the \cite{pmlr-v70-finn17a} experiments.
\subsection{Scenario} The goal and advantage of meta-learning is not only to achieve high performance within the trained distribution but also to generalize well to out-of-distribution datasets. To test this, we conducted two different scenario studies to see if performance could be maintained. The first scenario involves training on a general dataset and then conducting meta-test on a specific dataset \((G \rightarrow S)\). The second scenario involves training on a specific dataset and then conducting meta-test on a general dataset \((S \rightarrow G)\). For these experiments, we used the FC-100 dataset as the general dataset and the CUB dataset \cite{wah2011caltech} as the specific dataset.

\vspace{0.4cm}
\section{Experimental Results}

\begin{table*}[]
\centering
\begin{tabular}{r|cc|cc|cc|cc}
\toprule
Dataset & \multicolumn{2}{c|}{Omniglot} & \multicolumn{2}{c|}{Mini-ImageNet} & \multicolumn{2}{c|}{CUB} & \multicolumn{2}{c}{FC100} \\ 
\midrule
Num-Way & 5 & 20 & 3 & 5 & 3 & 5 & 3 & 5  \\ 
\midrule
MAML (1) ACC & 97.29 & 89.47 & 62.29 & 47.54 & 70.03 & 54.47 & 50.93 & 35.86 \\ 
\hfill CI  & 0.12 & 0.18 & 1.25 & 1.06 & 1.32 & 1.06 & 1.14 & 0.91\\ 
\midrule
AMFS (1) ACC & 98.16 & 91.87 & 63.43 & 47.83 & 69.47 & 56.32 & 50.35 & 36.17 \\ 
\hfill CI  & 0.10 & 0.18 & 1.29 & 0.95 & 1.45 &1.14 & 1.14 & 0.91  \\ 
\midrule
\midrule
MAML (5) ACC & 99.12 & 94.76 & 74.11 & 62.19 & 78.41 & 63.30 & 60.46 & 43.48 \\ 
\hfill CI  & 0.09 & 0.13 & 1.57 & 1.09 & 1.62 & 1.40 & 1.64 & 1.12\\ 
\midrule
AMFS (5) ACC & 99.38 & 96.10 & 74.37 & 64.14 & 78.29 & 64.46 & 59.96 & 43.77\\ 
\hfill CI  & 0.05 & 0.12 & 1.64 & 0.63 & 1.61 & 1.32 & 1.71 & 1.26\\ 
\bottomrule
\end{tabular}
\caption{Few-shot classification performance of local replication MAML and the AMFS framework. The CI shows 95\% confidence intervals over tasks. In the case of the Omniglot dataset, the accuracy is already very high even in the 5-way setting, so we conducted performance comparisons in a more challenging 20-way setting.}
\label{table:GBML_MAIN}
\end{table*}

The results in Table 1 compare our algorithm with MAML and First Order approximation MAML using the datasets mentioned earlier. We conducted meta-test with {3, 5}-way episode sampling. As shown in the Table 1, our algorithm outperformed the existing algorithms in most benchmarks. In lower-way settings, the variability of each episode can increase, making the gradient update process unstable, which can hinder model convergence and degrade performance. Despite this, our model demonstrated stable performance even in these low-way settings.

Traditional machine learning relies on the assumption that both training and testing data follow the same statistical patterns. However, in reality, unexpected distribution shifts can occur, causing model performance to drop when encountering unfamiliar data. This is known as the Out-Of-Distribution (OOD) problem. It is closely related to the key goal of meta-learning, which is ``learning to learn." Real-world applications are highly diverse and unpredictable, making it impossible to cover all possible scenarios with just the training data. Therefore, algorithms with strong OOD generalization capabilities can maintain performance even in unforeseen situations, making them effective in real-world environments. As a result, in our meta-learning approach, we evaluate our algorithm's domain generalization ability by conducting meta-tests on data that it has not seen before, a process we call ``scenario study." We validate generalization performance using two scenarios: training on a general dataset Mini-ImageNet, which includes various types of classes, and testing on a specific dataset CUB, and vice versa—training on a specific dataset and testing on a general dataset.We also use various ways and shots in these scenario studies and compare our results with the MAML algorithm. As shown in Table 2, our algorithm consistently outperformed the others, demonstrating the robustness and adaptability of our proposed method.

\subsection{Fast Convergence Capability of AMFS}
The MAML algorithm performs iterative gradient updates in its inner loop. These repetitive updates require substantial computation for each task, especially when the model has high capacity or the data is complex, often necessitating more steps. In contrast, our algorithm adapts to each task by performing a single closed-form optimization in the inner loop. This approach eliminates the need for repetitive computations, making it extremely fast and efficient.
Figure 2 compares the accuracy based on the number of gradient steps taken in the inner loop of the MAML algorithm versus our algorithm. Since I-AMFS always performs a single optimization step in the inner loop, it is represented by a fixed line. MAML's accuracy is compared by increasing the number of gradient steps in the inner loop from 1 to 5.
As shown in the figure, our algorithm achieves high performance without needing iterative optimization, while MAML's accuracy decreases with fewer gradient steps. The difference is particularly pronounced when only one gradient update is performed, highlighting the efficiency of our approach.

\begin{figure}
    \centering
    \includegraphics[width=8cm]{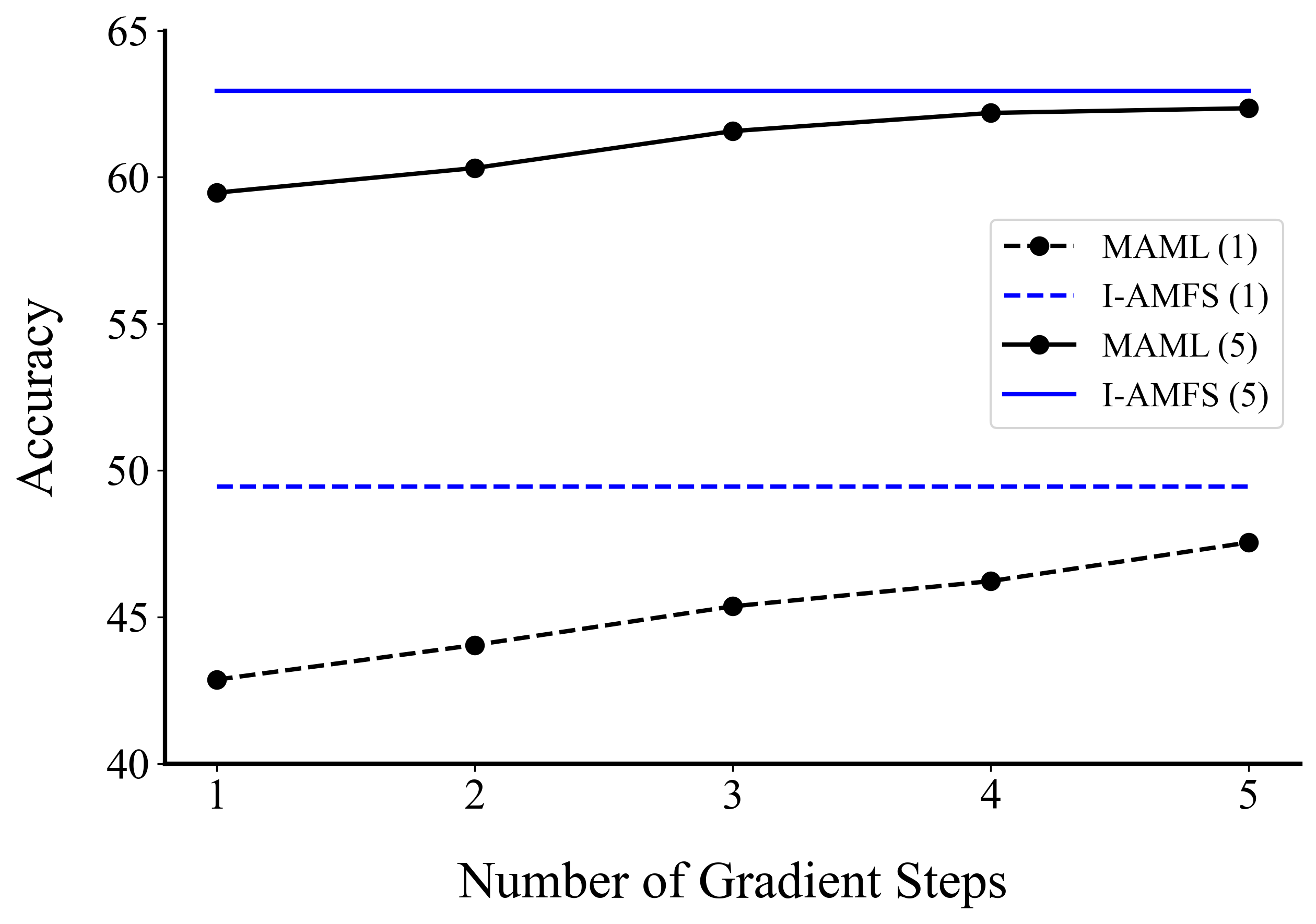}
    \caption{Accuracy based on the number of inner loop gradient updates for the Mini-ImageNet 5-way 1-shot and 5-shot tasks. Although the number of gradient update for I-AMFS was fixed at 1, it achieves faster convergence.}
    \label{fig:enter-label}
\end{figure}

Table 3 shows a comparison of our results with previous state-of-the-art methods in this task. The existing methods we compared include matching networks, meta-learner LSTM, and MAML, which were pioneers in this field. Our algorithm slightly outperformed these methods. Some of the previous approaches were designed specifically with few-shot classification in mind. However, our method, based on MAML, is independent of the specific task and can be applied not only to gradient-based algorithms but also to any algorithm that updates parameters in euclidean space.
Although all experiments were conducted using MAML as the baseline, the bi-level loops in our algorithm are orthogonal, meaning they can be integrated with any other meta-learning algorithm. Additionally, we plan to extend our algorithm from the euclidean parameter space to a non-euclidean geometric framework.

\begin{table}[]
\centering
\begin{tabular}{r|cc|cc}
\toprule
Scenario & \multicolumn{2}{c|}{G$\rightarrow{}$S} & \multicolumn{2}{c}{S$\rightarrow{}$G} \\ 
\midrule
Train $\rightarrow{}$ Test & \multicolumn{2}{c|}{Mini $\rightarrow{}$CUB} & \multicolumn{2}{c}{CUB$\rightarrow{}$Mini} \\ 
\midrule
Num-Way                   & 3 & 5 & 3 & 5 \\ 
\midrule
MAML (1) ACC  & 53.22 & 37.62 & 44.43 & 28.81  \\
\hfill CI  & 1.16 & 0.80 & 1.21 & 0.72 \\
\midrule
AFMS (1) ACC  & 53.64 & 38.65 & 43.85 & 30.10 \\
\hfill CI  & 1.25 & 0.79 & 1.07 & 0.74 \\
\midrule
\midrule
MAML (5) ACC  & 67.98 & 51.91 & 53.54 & 38.60 \\
\hfill CI  & 1.69 & 1.28 & 1.57 & 1.23 \\
\midrule
AFMS (5) ACC  & 66.92 & 54.92 & 54.07 & 37.19 \\
\hfill CI  & 1.75 & 0.63 & 1.47 & 1.12 \\
\bottomrule
\end{tabular}
\vspace{0.5cm}
\caption{Scenario test results. G: General, S: Specific}
\label{table:GBML_CROSS}
\end{table}

\begin{table}[]
\centering
\begin{tabular}{r|cc}
\toprule
Datasets & \multicolumn{2}{c}{Mini-ImageNet} \\ 
\midrule
Num-Way & \multicolumn{2}{c}{5} \\ 
\midrule
Num-Shot                   & \multicolumn{1}{c|}{1} & 5  \\ 
\midrule
fine-tuning baseline  &  \multicolumn{1}{c|}{28.86 $\pm$ 0.54\%} & 49.79 $\pm$ 0.79\% \\
\midrule
matching nets & \multicolumn{1}{c|}{43.56 $\pm$ 0.84\%} & 55.31 $\pm$ 0.65\% \\
\midrule
meta-learner LSTM & \multicolumn{1}{c|}{43.44 $\pm$ 0.77\%} & 60.60 $\pm$ 0.71\% \\
\midrule
FO-MAML & \multicolumn{1}{c|}{48.07 $\pm$ 1.75\%} & 63.15 $\pm$ 0.91\% \\
\midrule
MAML & \multicolumn{1}{c|}{48.70 $\pm$ 1.84\%} & 63.11 $\pm$ 0.92\% \\
\midrule
* FO-MAML & \multicolumn{1}{c|}{44.22 $\pm$ 1.12\%} & 58.86 $\pm$ 1.15\% \\
\midrule
* MAML & \multicolumn{1}{c|}{47.54 $\pm$ 1.06\%} & 62.19 $\pm$ 1.09\% \\
\midrule
AMFS & \multicolumn{1}{c|}{\textbf{48.83} $\pm$ 0.95\%} & \textbf{64.14} $\pm$ 0.63\% \\
\bottomrule
\end{tabular}
\vspace{0.5cm}
\caption{Performance comparison with previous researches. Results of this table not otherwise indicated are from \cite{pmlr-v70-finn17a}, \cite{vinyals2016matching}. *: local replication}
\vspace{-0.3cm}
\label{table:GBML_CROSS}
\end{table}

\vspace{0.4cm}
\section{Discussion}
Our analysis of the experimental results shows that the AMFS framework generally achieves higher accuracy in few-shot classification compared to MAML, though the margin of improvement is modest in most cases. The primary advantage of AMFS is its fast convergence, especially in the inner loop, where the kernel-based approach promotes both stability and speed. In the outer loop, the adaptive updating of meta-parameters helps the meta-learner learn more efficiently.
Another advantage of AMFS is its ability to handle the challenges of First-Order approximation. As you can see in Table 4, while MAML experiences significant performance deterioration when using First-Order approximation for certain tasks, AMFS shows more resilience. According to \cite{nichol2018first}, the performance of First-Order MAML (FO-MAML) can be improved by adjusting the meta-learning rate. The observed results suggest that O-AMFS indirectly helps adjust the meta-learning rate through its process of determining the meta-gradient, reducing the negative impact of First-Order approximation. This finding is significant because it suggests the potential to automatically and appropriately adjust the hyperparameters of MAML, which are typically highly sensitive to tuning, particularly with respect to the meta-learning rate. \cite{antoniou2019train}

Nevertheless, there are areas where the AMFS framework could be improved. First, the weight allocation method in O-AMFS is relatively simple. At present, the weights assigned to the loss of each task are determined based on the similarity of the gradients $\theta_i^\prime$ obtained from each task $\mathcal T_i$ within the inner loop. Further research is required to explore various modifications aimed at identifying optimal criteria for measuring gradient similarity and methods for weight allocation.

Another consideration is that for the similarity of task gradients to meaningfully influence the outer loop updates, a sufficient number of tasks-or equivalently, a large enough batch size-is necessary. However, we followed the experimental setup from \cite{pmlr-v70-finn17a}, where the batch size is typically 2 or 4, which may limit the influence of gradient similarity. Consequently, future research should investigate the optimal trade-off between computational efficiency and performance with respect to batch size.

While the ideas within the AMFS framework are still in the early stages of development, they address important challenges in meta-learning. Meta-learning, at its core, is a field dedicated to developing methods that ensure fast learning and generalized performance, particularly in scenarios where train data or computational resources are limited. However, the process of sampling tasks during adaptation increases memory usage proportional to the number of tasks. Moreover, in an $n$-way $k$-shot learning setup, the amount of data actually required for meta-learner training far exceeds the original intent of meta-learning, which aims for efficiency.

In contrast, the AMFS framework shows superior data efficiency, as the data used in the function space during the inner loop depends on $nk$ rather than $k$. Additionally, the inherent rapid convergence of the AMFS framework contributes to addressing memory efficiency challenges.

As deep learning models continue to grow in scale and the required amount of training data expands accordingly, resource constraints are increasingly raising the barriers to entry in research. For the field of artificial intelligence to advance in a desirable manner, these barriers need to be lowered, and this necessitates a renewed focus on fulfilling the original goals of meta-learning. This paper raises fundamental questions about the core principles of meta-learning and aims to lay the groundwork for returning to its foundational objectives.

\begin{table}[]
\centering
\begin{tabular}{r|cc|cc}
\toprule
Datasets & \multicolumn{4}{c}{Mini-ImageNet} \\ 
\midrule
Num-Way    & \multicolumn{2}{c|}{3} & \multicolumn{2}{c}{5} \\ 
\midrule
Num-shot  & 1 & 5 & 1 & 5 \\ 
\midrule
MAML ACC  &  62.29 & 74.11 & 47.54 & 62.19 \\
\hfill CI  & 1.25 & 1.57 & 1.06 & 1.09 \\
\midrule
FO-MAML ACC  &  \textbf{44.07} & \textbf{52.68} & 44.22 & 58.86 \\
\hfill CI  &  0.97 & 1.65 & 1.12 & 1.15 \\
\toprule
AMFS ACC  &  63.43 & 74.37 & 48.83 & 64.14 \\
\hfill CI  & 1.29 & 1.64 & 0.95 & 0.63 \\
\midrule
FO-AMFS ACC  &  60.01 & 67.80 & 44.66 & 59.16 \\
\hfill CI  &  1.23 & 1.71 & 1.01 & 1.13 \\
\bottomrule
\end{tabular}
\vspace{0.5cm}
\caption{When applying First-Order Approximation to both the MAML and AMFS frameworks, MAML exhibited a noticeable performance drop in some tasks, while AMFS remained relatively robust. (FO: First-Order Approximation)}
\label{table:GBML_CROSS}
\end{table}

\vspace{0.4cm}
\section{Conclusion}
In this paper, we have introduced two novel algorithms designed to address the inherent limitations of the Model-Agnostic Meta-Learning (MAML) framework, focusing on improving both its computational efficiency and stability. Our first algorithm redefines the optimization process within the function space, allowing for model updates using closed-form solutions rather than the traditional, computationally expensive gradient steps in the inner loop. The second algorithm enhances the outer loop by dynamically weighting the losses from each task. Defined by these two algorithms, AMFS framework optimizes the convergence process during both training and inference stages.

Through extensive experiment, we demonstrated that our proposed AMFS framework not only achieves higher few-shot classification accuracy compared to the original MAML but also exhibits faster convergence and greater stability, particularly in the inner loop. The resilience of AMFS to the negative effects of First-Order approximation further underscores its robustness, making it a more reliable option for meta-learning tasks.

However, while our results are promising, they also highlight areas for future research. The weight allocation method in the outer loop, though effective, remains relatively simplistic and could benefit from further refinement to optimize task-specific gradient similarity measures. Additionally, exploring the balance between batch size and computational efficiency will be crucial for fully leveraging the advantages of the AMFS framework.

Ultimately, this work contributes to the ongoing discourse on the essence of meta-learning, challenging conventional approaches and suggesting a return to its foundational objectives—rapid learning and efficient generalization with limited data and resources. As deep learning continues to evolve, the need for more efficient, adaptable models becomes increasingly critical. Our findings lay the groundwork for a new paradigm in meta-learning, one that prioritizes both theoretical rigor and practical applicability in real-world scenarios.


\vspace{0.4cm}
\bibliographystyle{unsrt}  
\bibliography{references}

\end{document}